# Sensing Population Distribution from Satellite Imagery via Deep Learning: Model Selection, Neighboring Effects, and Systematic Biases

Xiao Huang, Di Zhu, Fan Zhang, Tao Liu, Xiao Li, and Lei Zou.

*Abstract*—The rapid development of remote sensing techniques provides rich, large-coverage, and high-temporal information of the ground, which can be coupled with the emerging deep learning approaches that enable latent features and hidden geographical patterns to be extracted. This study marks the first attempt to cross-compare performances of popular state-of-the-art deep learning models in estimating population distribution from remote sensing images, investigate the contribution of neighboring effect, and explore the potential systematic population estimation biases. We conduct an end-to-end training of four popular deep learning architectures, i.e., VGG, ResNet, Xception, and DenseNet, by establishing a mapping between Sentinel-2 image patches and their corresponding population count from the LandScan population grid. The results reveal that DenseNet outperforms the other three models, while VGG has the worst performances in all evaluating metrics under all selected neighboring scenarios. As for the neighboring effect, contradicting existing studies, our results suggest that the increase of neighboring sizes leads to reduced population estimation performance, which is found universal for all four selected models in all evaluating metrics. In addition, there exists a notable, universal bias that all selected deep learning models tend to overestimate sparsely populated image patches and underestimate densely populated image patches, regardless of neighboring sizes. The methodological, experimental, and contextual knowledge this study provides is expected to benefit a wide range of future studies that estimate population distribution via remote sensing imagery.

## I. INTRODUCTION

FINE knowledge of the spatial contribution of human activity is essential for a wide range of fields, such as public health [1]–[3], urban planning [4]–[6], disaster management [7], [8], resource allocation [9], economic evaluation [10], and migration [11], [12]. As stated by the global sustainable development goals (SDGs), understanding where and how people are distributed is of great importance to make "cities and human settlements inclusive, safe, resilient, and sustainable".

Census data, when linked with accurate administrative boundary data, can provide spatially explicit population distribution. In the U.S., for instance, official population data on various geographical levels (e.g., the Decennial Census records and the American Community Survey (ACS)) are repetitively released by the U.S. Census Bureau. Similar agencies that regularly release population distribution data include the Office for National Statistics of the U.K., the National Bureau of Statistics of China, and the Statistics Bureau of Japan, to list a few. Despite the authority of census-based population distribution released by the officials, it owns several intrinsic limitations, making it ill-suitable for many spatial problems. First, population distribution is with great heterogeneity [13]; therefore, it can not be assigned uniformly distributed within predefined geographical units. Second, the census-based population suffers from the Modifiable Areal Unit Problem (MAUP) [14] due to its arbitrarily imposed boundaries that are rarely consistent with other boundaries in practical applications [15]. Third, census-based population distribution is often with poor temporal resolutions that preclude temporal-dynamic population estimations, and recent and reliable population data at fine scales can often be lacking, especially in resource-poor settings [16], [17]. Given the above limitations, scholars start to explore various means to improve the aggregated census-based population, one notable effort of which is to derive fine-grained, spatially-continuous population grids.

A population grid refers to a geographically referenced lattice of square cells, with the value of each cell representing a population count at its location. Population grids are generally constructed based on census unit-based data via dasymetric modeling [18], [19], or other statistical approaches that intelligently assign population to grids by establishing relationships between population and supporting auxiliary variables [20], [21]. The derived population grids not only capture the heterogeneity of population distribution but also

Xiao Huang is with Department of Geosciences, University of Arkansas, e-mail: xh010@uark.edu (Corresponding author).

Di Zhu is with Department of Geography, Environment, and Society, University of Minnesota, e-mail: dizhu@umn.edu.

Fan Zhang is with Department of Urban Studies and Planning, Massachusetts Institute of Technology, e-mail: zhangfan@mit.edu.

Tao Liu is with College of Forest Resources and Environmental Science, Michigan Technological University, e-mail: taoliu@mtu.edu.

Xiao Li is with Texas A&M Transportation Institute, Texas A&M University, e-mail: xiao.li@tamu.edu.

Lei Zou is with Department of Geography, Texas A&M University, e-mail: lzou@exchange.tamu.edu.

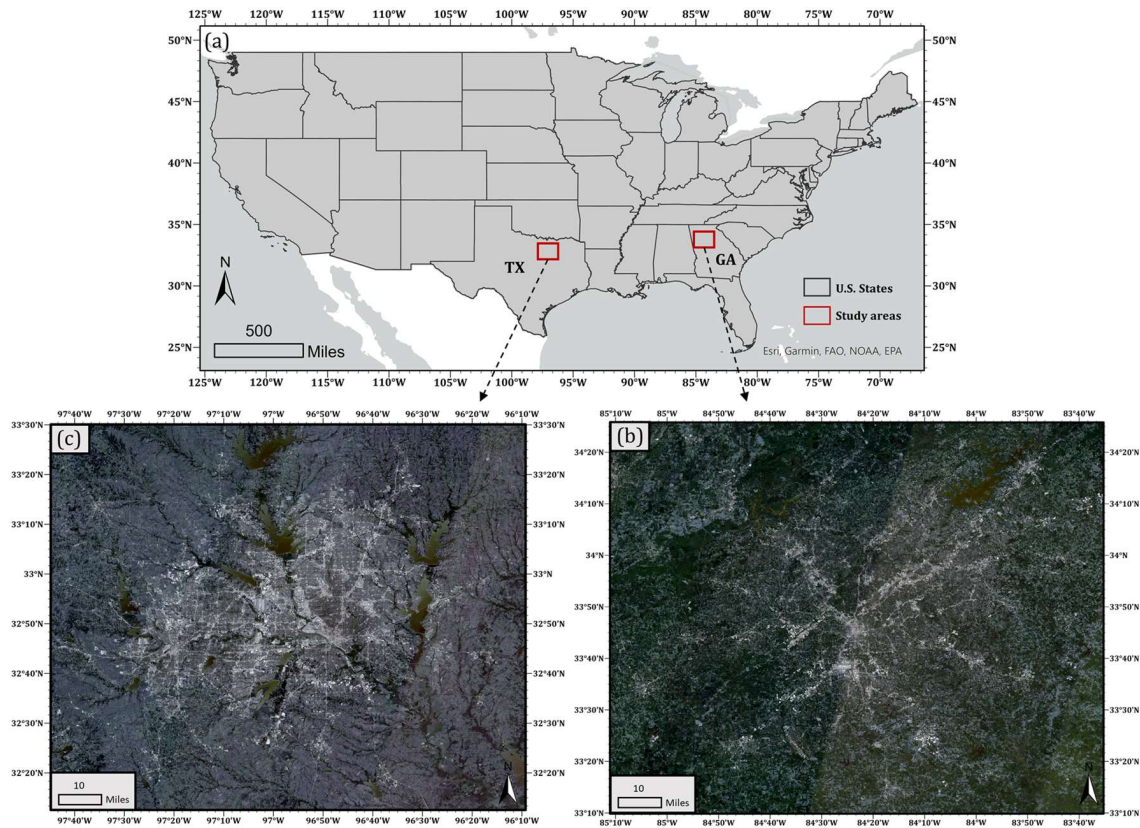

Fig. 1. Study areas with two sites. (a) Conterminous U.S.; (b) Metro Atlanta (model training, validating, and testing); (c) Metro Dallas (evaluating generalizability).

ensure the aggregated numbers at census units match the official records [22]. Despite that population distribution delivered in fine-grained gridded format achieves spatial heterogeneity and largely mitigates the MAUP, the production of accurate population grids is temporally-restricted to the release of census data [16] and is heavily dependent on the availability of a diverse set of inputs (often large in number), leading to its questionable sustainability. With the development of statistical modeling techniques, questions arise: can we obtain a mapping between the gridded population and the gridded latent variables? If so, can we achieve a relatively robust mapping using as few and as easily-accessible variables as possible?

The rapid development of remote sensing techniques facilities the consistency and sustainability of auxiliary variables, which are often preferred and needed in population distribution modeling, as remote sensing imagery, with improving imaging capability over time, provides rich, large-coverage, and high-temporal information of the ground. The advantages of remote sensing imagery are further expended by the emerging supervised deep learning approaches with the capacity of extracting both low- and higher-level latent features and capturing hidden hierarchies of geographical patterns from images [23]–[25], thus forming a stable supervised end-to-end mapping [26]. The coupling of remote sensing imagery and deep learning algorithms undoubtedly establishes a new venue that potentially advances traditional population modeling. Numerous efforts have been made to harvest the strong mapping capability of deep learning in estimating both population count and population density directly from satellite images [27]–[30]. These efforts differ in architecture design, model selection, and evaluation metrics. Despite the aforementioned attempts, gaps still exist in 1) evaluating the performance of state-of-the-art deep learning models to provide guidelines of model selection for future studies, 2) thoroughly investigating the potential contribution of neighboring image patches to the population estimation in the center image patch (neighboring effects) to provide guidance for the selection of suitable neighboring sizes, and 3) exploring the potential systematic population estimation biases resulting from the intrinsic limitations of remote sensing images.

To fill the above gaps, we perform end-to-end training on popular deep learning architectures by establishing a mapping between satellite image patches and their corresponding population count from an existing gridded population product. Taking the Metropolitan Atlanta (Metro Atlanta) and Metropolitan Dallas (Metro Dallas) as the study region, we obtain remote sensing images from Sentinel-2 with a spatial resolution of 10m and derive the 24-hour ambient gridded population distribution from LandScan (https://landscan.ornl.gov/). Models are trained, validated, and tested in Metro Atlanta and further applied to Metro Dallas to evaluate the overall generalizability. The main contributions of this article are summarized as follows.

1) We adopt transfer learning techniques by modifying and fine-tuning several popular deep learning architectures

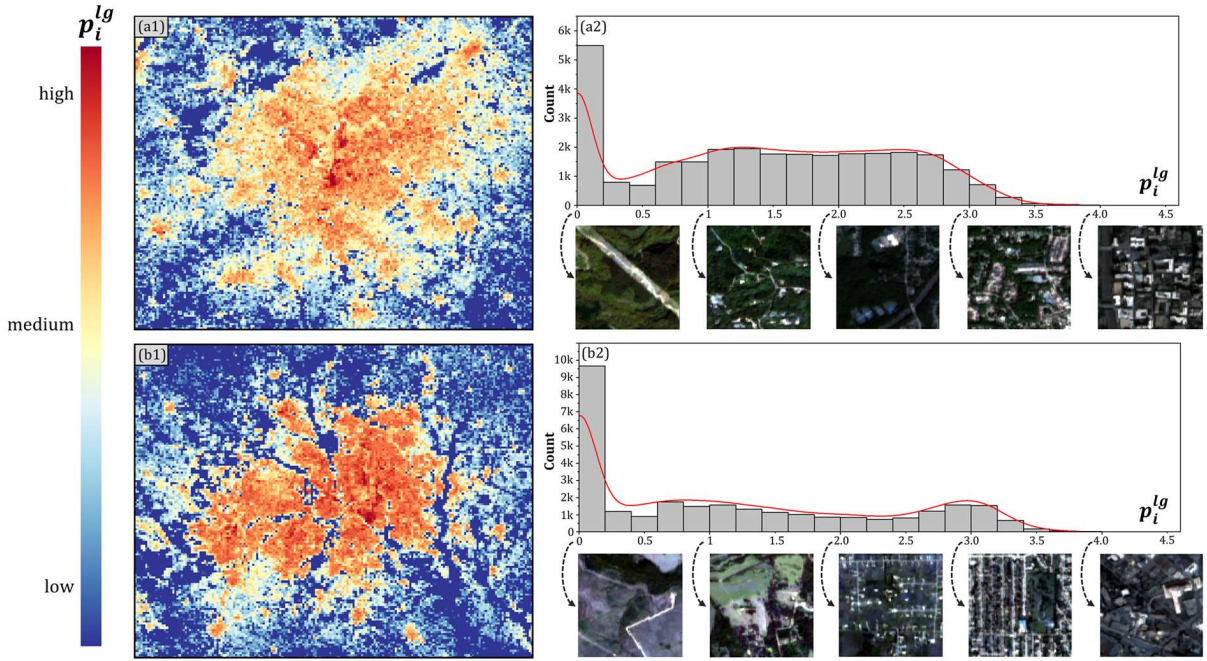

Fig. 2. *The distribution of the logarithmized population within each grid ($p_i^{lg}$) in Metro Atlanta (a1) and Metro Dallas (b1). Histograms and examples of image patches corresponding to the logarithmized population in Metro Atlanta (a2) and Metro Dallas (b2).*

towards the population estimation task. We further cross-compare estimated population distribution from these architectures quantitatively and qualitatively, providing guidelines for the model selection in future studies.

2) We investigate the neighboring effects by progressively extending the selection of neighboring patches to estimate the population residing in the center image patch, offering guidance for the selection of suitable neighboring sizes.

3) We explore the potential systematic biases in population estimation directly from remote sensing images via deep learning approaches, discuss reasonable causes that lead to these biases, and provide applicable solutions to mitigate these biases.

## II. RELATED WORK

### A. Existing population grid products and their issues

Gridded population contain regulated cells with population count, serving as an ideal input for training a mapping between image patches and the corresponding population statistics. Numerous global and regional-focused population grid products have been developed and released with various spatiotemporal scales. Popular population products that contain the U.S. include (but is not limited to): Gridded Population of the World (GPW) [31], Global Rural Urban Mapping Project (GRUMP) [21], Global Human Settlement–Population (GHS-POP) [32], World Population Estimate (WPE) [33], LandScan-USA [34], and Building-based Population Grid USA (BPG-USA) [13]. Among them, GPW, GRUMP, GHS-POP, and WPE are at a global scale, while LandScan-USA and BPG-USA specifically target the U.S. and Conterminous U.S., respectively. Dasymetric techniques are often applied to allocate population count to habitable cells within each census unit, but with different input of auxiliary variables and weighting scenarios. Commonly used auxiliary variables involve land use/land cover (usually derived from remote sensing imagery), nighttime light intensity, distribution of infrastructures (e.g., roads, buildings, and point of interest), environmental variables, and restrictions (e.g., water body and protected areas), to list a few. Population grid products generally ensure that aggregated numbers at census units match the official records, at the same time, capture the heterogeneity of population distribution using regulated cells. However, massive inputs, especially for accurately modeled gridded population, need to be prepared and updated in accordance with the release of the new census population [35], posing a great challenge to the continuous product release, as auxiliary variables with differing spatiotemporal scales might be difficult to collect [16]. In addition, census-disaggregated population grids largely rely on the availability of census data. Such an issue is exaggerated in data-poor settings, where census data and required auxiliary variables are obsolete or unavailable at all.

The development of remote sensing platforms with improving imaging capability leads to the easy acquisition of rich, large-coverage, and high-temporal information of the ground, serving as an ideal set of variables that benefits continuous population distribution mapping. Deep learning techniques further enhance the ability to extract high-level latent features with hidden hierarchies of geographical patterns. Once a stable mapping is established between the gridded population and the corresponding image patch, population distribution mapping can be achieved only with remote sensing images, which are easily accessible and regularly updatable.

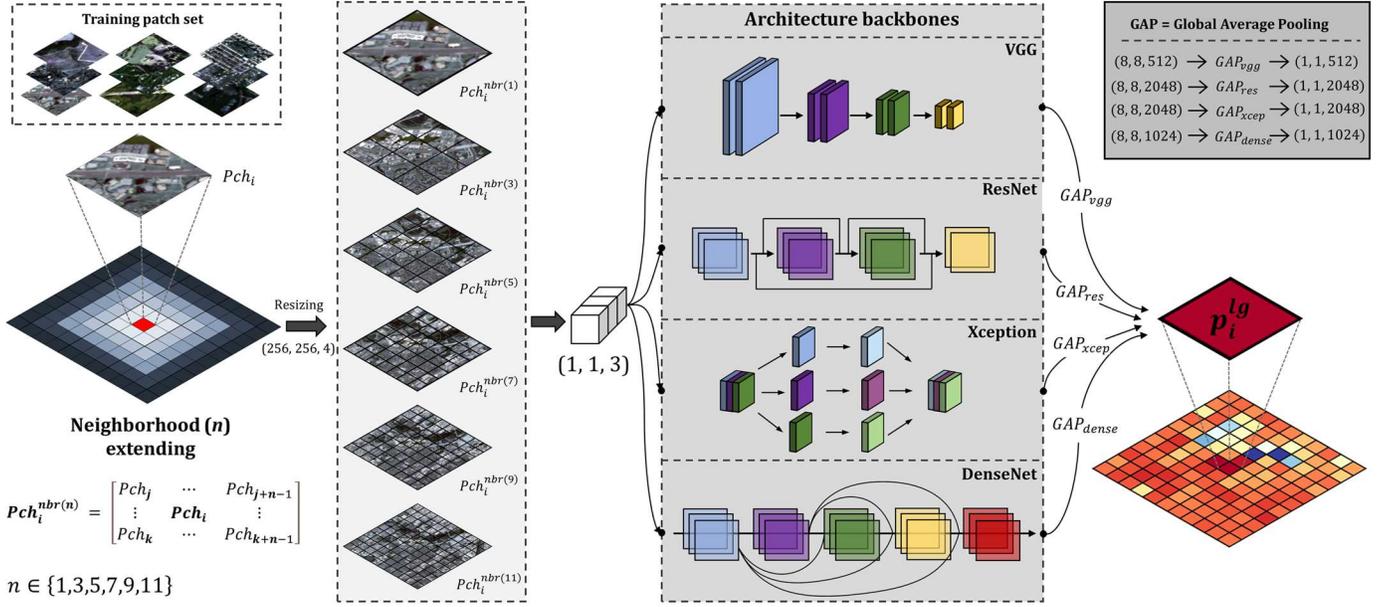

Fig. 3. An end-to-end framework to derive the mapping between remote sensing imagery and population distribution. This framework tests the performances of popular deep learning architectures with different neighboring considerations. A dropout ratio of 0.5 is implemented after the results from the global average pooling layers are flattened. Note that $Pch_i^{nbr(n)}$ contains four channels, i.e., R, G, B, and NIR.

## B. Popular deep learning architectures and transfer learning

Most modern deep learning models are based on artificial neural networks, specifically convolutional neural networks (CNN) that were inspired by the visual system's structure and hierarchically composed of input, output, and multiple hidden layers [36]. AlexNet by Krizhevsky et al. [37] was a pioneering CNN that transcended traditional approaches in various computer vision tasks. Further, the VGG by Simonyan and Zisserman [38] improved the performance by stacking simple convolutional operations, forming a deep architecture. Diverging from the mainstream of stacking layers via a sequential structure, GoogLeNet [39] achieved great performance by introducing a building block, i.e., the inception module, that largely reduces the number of parameters and operations. Numerous updated versions of GoogLeNet were proposed: Inception-V2 and -V3 [40], Inception-V4 [41], and Xception [42]. ResNet [43] is well-known for its depths and the introduction of residual blocks that implement identity skip connections to promote gradient propagation. From another perspective in solving the problem of vanishing gradient, DenseNet [44] facilitates the acquisition of "collective knowledge" by allowing each layer to obtain additional inputs from all preceding layers.

Training the above architectures from scratch is not often feasible due to the limited sample size and the temporal inefficiency for models with randomly initialized parameters to converge. Studies have proved that using pre-trained weights from similar tasks greatly benefits the model training process [45], [46]. In this study, we adopt transfer learning techniques by loading the weights pre-trained from ImageNet (http://www.image-net.org/) as initialization. We select four popular deep learning architectures and compare their performances in deducing population counts given the corresponding remote sensing image patches. These modern architectures, including VGG, ResNet, Xception, and DenseNet, are widely applied in various computer vision tasks and generally diverge from each other in the overall architecture design.

## C. Deriving population distribution directly from satellite imagery via deep learning

Numerous attempts have been made to derive population distribution from satellite imagery supported by the strong capability of feature extraction in deep learning. Doupe et al. [30] converted satellite images into population density estimates using a VGG-like architecture that includes collections of "Convolutional Layer (Conv), Pooling Layer (Pool) and Rectified Linear Unit (ReLU)" followed by two Fully-Connected Layers (FC), each with 4096 neurons. They trained and implemented their model in Tanzanian and Kenyan and found that their model achieved great performance with decent generalizability. Robinson et al. [29] adopted a similar VGG-like sequential architectural design to Doupe et al. [30] but regarded the task as a classification problem instead of a regression problem. They aimed to classify image patches to labels that derive from the power level of the population count (14 classes in total). The results suggested that the predictions from their model were consistent with the ground-truthing labels with an $R^2$ over 0.9 in selected areas in the U.S. Hu et al. [27] built a customized CNN architecture to predict population density in India from multisource imagery fused by the implementation of $1 \times 1$ Conv layers. Their results, again, demonstrated the feasibility of producing accurate population estimates directly from satellite imagery, especially for rural areas. A more recent effort was by Xing et al. [28], who implemented a ResNet-based architecture with the consideration of neighboring effects, termed as Neighbor-

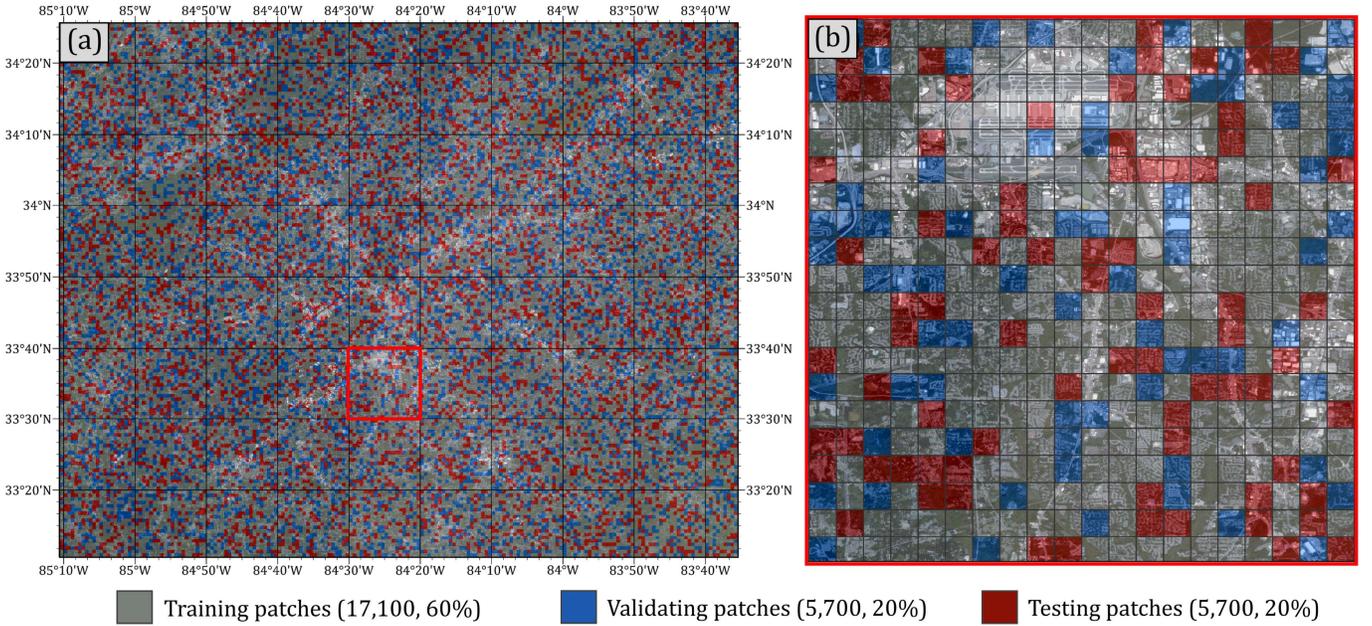

*Fig. 4. The spatial distribution of Metro Atlanta's image patches in the training set (60%), validating set (20%), and testing set (20%).*

**Training patches (17,100, 60%)**    **Validating patches (5,700, 20%)**    **Testing patches (5,700, 20%)**

ResNet. The proposed Neighbor-ResNet architecture used ResNet as a backbone and aimed to estimate the population residing in the center image patch, with the support of the eight surrounding neighboring patches (within a 3 × 3 patch space). The results proved the superiority of Neighbor-ResNet compared to regular ResNet that ignores the neighboring effects.

Despite the above efforts, few studies conduct comparisons among state-of-the-art deep learning architectures in terms of evaluating their performances in the mapping between satellite imagery and population. The recent effort by Xing et al. [28] identified the positive contribution of neighboring patches but with a fixed 3 × 3 patch space. Efforts are needed to explore the dynamics of performance when a different size of a surrounding region is considered.

## III. DATASETS AND STUDY AREAS

### A. Datasets

#### 1) Population grid (LandScan USA Population Database)

In this study, we selected the population grid from LandScan (https://landscan.ornl.gov/) as ground-truthing population distribution, serving as a reference that allows an end-to-end mapping to be established by deep learning architectures between satellite patches and the corresponding population count. LandScan belongs to Oak Ridge National Laboratory (ORNL), the largest science and energy national laboratory in the Department of Energy (DOE). LandScan's population grid has been widely recognized as one of the community standard products [47] and applied in a variety of domains [48]–[50]. LandScan team adopts multivariate dasymetric modeling frameworks using the best available demographic (Census) data, remote sensing imagery, and other supporting variables to disaggregate census population counts into grids [34]. The population product we selected is the LandScan USA Population Database 2019, currently hosted by Homeland

Infrastructure Foundation-Level Data (HIFLD). It provides estimated population counts at 3 arc-second resolution for Nighttime and Daytime for Conterminous U.S., Hawaii, Alaska, as well as other U.S. territories. We averaged the Daytime and Nighttime population grids to form a 24-hour ambient population grid. Note that LandScan USA Population Database 2019 is a static baseline population estimate, which does not include transitory populations such as business travelers and tourists [51]. Given the fact that the population grid with a 3 arc-second resolution presents great heterogeneity that medium-resolution satellite imagery might fail to capture (as a 3 arc-second grid only contains a limited number of pixels in the imagery), we aggregated the original population grid to 30 arc-second (approximately 1 km at the equator), a spatial resolution adopted by many studies similar to this work [27]–[29].

#### 2) Satellite imagery (Sentinel-2)

The satellite imagery used in this study was derived from Sentinel-2, a wide-swath, fine-resolution, multispectral imaging mission of the European Space Agency (ESA) developed in the framework of the European Union Copernicus program [52]. We selected the Sentinel-2 Level-2A product that had been atmospherically corrected (bottom of atmosphere reflectance) and orthorectified. We retrieved four bands, i.e., Red (R), Green (G), Blue (B), and Near-infrared band (NIR), all with a spatial resolution of 10 m. Studies have demonstrated that the physical and chemical characteristics of various types of land use and land cover can be well reflected by these four bands [53], [54]. Thus, we expected a stable mapping that associates population count with these bands to be formed by deep learning architectures through end-to-end training. In addition, NIR and visible spectrum that includes R, G, B are available in most multispectral sensors. Serving as model inputs, the easy accessibility of these bands greatly promotes generalizability and sustainability. We queried Sentinel-2 imagery via Google Earth Engine (GEE). The imagery covers

Metro Atlanta and Metro Dallas (details in Section 3.2) with a temporal coverage from January 1 to December 31, 2019, consistent with the time span of the 2019 LandScan population grid. We implemented a standard Sentinel-2 cloud mask in GEE and selected the median value for each pixel with overlapping values among different scenes.

### B. Study areas

We selected two metropolitan regions within the Conterminous U.S. (Fig. 1a) as our study areas, i.e., Metro Atlanta (Fig. 1b) and 2) Metro Dallas (Fig. 1c). The study site that covers Metro Atlanta is bounded with latitude from $33.1750° N$ to $34.4250° N$, and longitude from $83.5916° W$ to $85.1750° W$. The study site that covers Metro Dallas is bounded with latitude from $32.2325° N$ to $33.4825° N$, and longitude from $96.1500° W$ to $97.7334° W$. These two metropolitan areas are both densely populated metroplex but with observable discrepancies in their distribution patterns of land use and land cover. Metro Atlanta is characterized by its sprawl-out urban fabrics (Fig. 1b). In comparison, the urban fabrics in Metro Dallas are comparably centralized, evidenced by its distinctly dense urban core (Fig. 1c). We further divided the two study sites into 30 arc-second grids, leading to a total of $28,500$ ($190 \times 150$) grids in each site for model training purposes. We trained deep learning architectures using patches from Metro Atalanta and evaluated their generalization capability using patches from Metro Dallas. We believe that the dissimilarity between these two sites benefits us in observing potential overfitting issues and systematic biases.

## IV. METHODOLOGY

### A. Preprocessing and problem formation

We first resized the daytime and nighttime population grids from LandScan to 30 arc-second and averaged them to form 24-hour ambient population grids. Let $p_i^{day}$ and $p_i^{night}$ respectively denote the daytime and nighttime population count in grid $i$, the 24-hour ambient population count in grid $i$ ($p_i$) is calculated as

$$p_i = \begin{cases} \frac{p_i^{day} + p_i^{night}}{2} & , if \ \frac{p_i^{day} + p_i^{night}}{2} \geq 1 \\ 0 & , else \end{cases} \quad (1)$$

For each grid $i$, there exists a corresponding image patch $i$, denoted as $Pch_i$. To investigate the potential contribution of neighboring patches to the population residing in the center image patch, we extended $Pch_i$ to include its neighboring patches. Given image patch $Pch_i$, the extension, i.e., $Pch_i^{nbr(n)}$, that includes its $n \times n$ neighborhood can be formulated as

$$Pch_i^{nbr(n)} = \begin{bmatrix} Pch_j & \cdots & Pch_{j+n-1} \\ \vdots & Pch_i & \vdots \\ Pch_k & \cdots & Pch_{k+n-1} \end{bmatrix} \quad (2)$$

where $n$ denotes the size of the neighborhood and $n \in \{1,3,5,7,9,11\}$.

As we observed that the distribution of $p_i$ was heavily tailed (even excluding the value of 0), we took the logarithms of the population count at grid $i$, denoted as $log p_i$, following the works by Xing et al. [28] and Hu et al. [27]: $p_i^{lg} = log_{10}(p_i)$. Specifically, if $p_i = 0$, we set $p_i^{lg}$ as 0. Figure 2 presents the distribution of $log p_i$ in the two study areas: Metro Atlanta (Fig. 2a1) and Metro Dallas (Fig. 2b1). Their histograms and examples of image patches corresponding to the logarithmized population count are presented in Fig. 2a2 and Fig. 2b2, respectively. We observed a spike of 0 values in both study areas, suggesting the existence of massive uninhabited grids. The logarithm operation flattened the curve for values above 0, leading to a considerably balanced target set that benefits model training.

We view population distribution estimation as a mapping problem. Given the extended image patch centered at $Pch_i$, i.e., $Pch_i^{nbr(n)}$, and the logarithmized population count at grid $i$, i.e., $log p_i$, we aim to build a mapping function between them:

$$Pch_i^{nbr(n)} \rightarrow p_i^{lg} \quad (3)$$

where symbol $\rightarrow$ denotes the mapping to be learned by deep learning architectures. Note that $Pch_i^{nbr(n)}$ contains four channels, i.e., R, G, B, and NIR. We detailed this end-to-end architecture in the next session.

### B. End-to-end framework

We implemented an end-to-end training framework to facilitate the establishment of the mapping from $Pch_i^{nbr(n)}$ to $p_i^{lg}$ (Figure 3). Considering the dynamic yet elusive relationship between remotely observed imagery and population distribution, deep neural networks are often regarded as preferred choices in modeling such a nonlinear, complex relationship [28]. In this study, we selected four widely-adopted deep learning architectures with diverging concepts in design: VGG (VGG-16), ResNet (ResNet-50), Xception, and DenseNet (DenseNet-121). We aim to compare their performances in deducing population count given the corresponding image patches with different neighborhood considerations. As shown in Fig. 3, we first reconstructed $Pch_i^{nbr(n)}$ by considering the $n$ by $n$ neighborhood of $Pch_i$. We then resized $Pch_i^{nbr(n)}$ to a fixed size of $256 \times 256$. Each reconstructed $Pch_i^{nbr(n)}$ has four channels, i.e., R, G, B, and NIR. Aiming to fine-tune weights of backbone models pre-trained from ImageNet, we implemented $1 \times 1$ Conv layers, commonly-used layers for image fusion and dimensionality reduction [43], to reduce the number of channels from four to three, consistent with the situation where the initial weights from ImageNet were derived. Further, we fed the re-channelized $Pch_i^{nbr(n)}$ to the backbones (with top fully connected layers cut off) of the four selected deep learning architectures. We appended a global average pooling layer to each architecture backbone, flattened the resulted neurons with 0.5 as dropout ratio before connecting them to the output neuron, whose linear activation suggests the estimated $p_i^{lg}$, denoted as $\widehat{p_i^{lg}}$.

# Atlanta

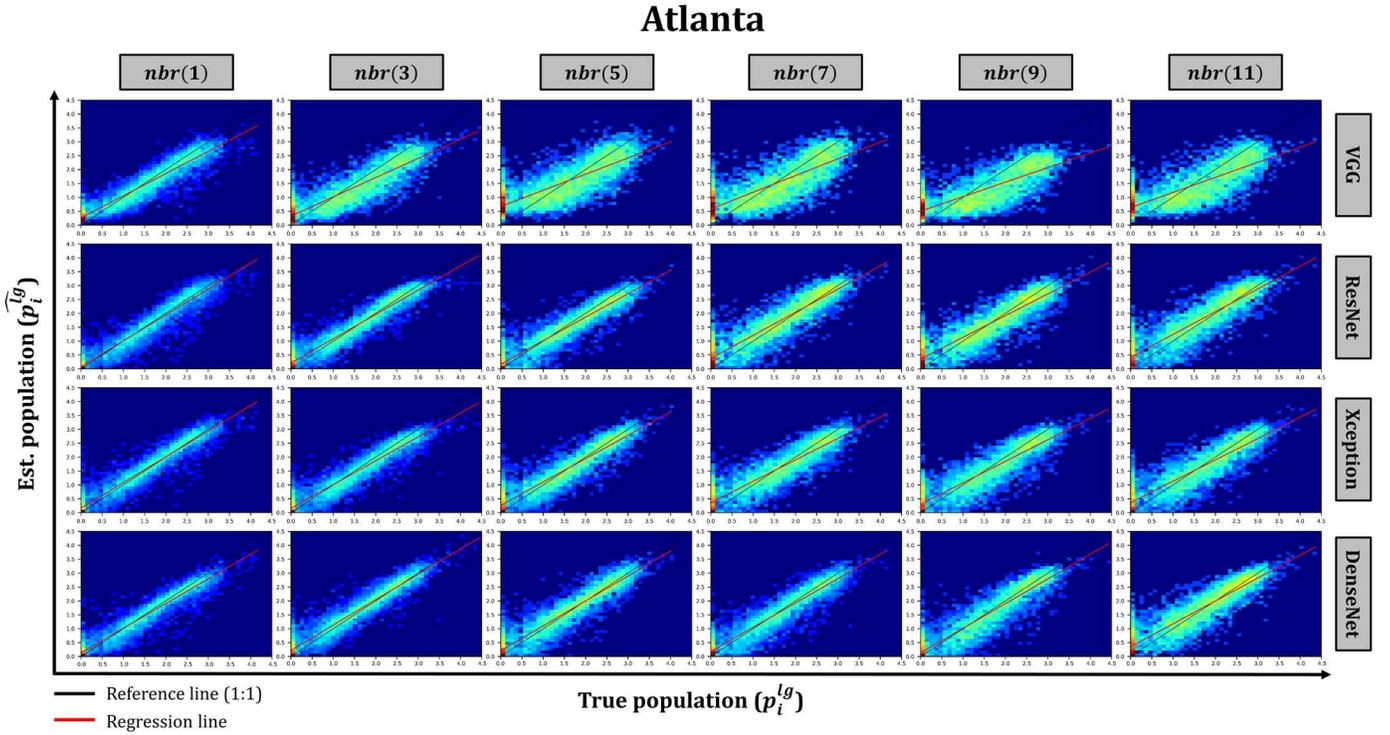

**Fig. 5.** *Scatterplot of Metro Atlanta testing patches between the true population ($p_i^{lg}$) and the estimated population ($\widehat{p_i^{lg}}$) from VGG, ResNet, Xception, and DenseNet, under different neighboring scenarios.*

## C. Training strategies

We chose log hyperbolic cosine (log-cosh) loss for back-propagation learning and weight updating:

$$\mathcal{L}\left(p_i^{lg}, \widehat{p_i^{lg}}\right) = \sum_{i=1}^{s} log_{10}\left(cosh\left(\widehat{p_i^{lg}} - p_i^{lg}\right)\right) \quad (4)$$

where $\mathcal{L}\left(p_i^{lg}, \widehat{p_i^{lg}}\right)$ denotes the log-cosh loss function, given the ground-truthing $p_i^{lg}$ and estimated $\widehat{p_i^{lg}}$, $s$ demotes the batch size (set as 32), and $cosh\ (x)$ denotes the hyperbolic cosine function, i.e., $\frac{e^x + e^{-x}}{2}$. The log-cosh is similar to the mean squared error loss (i.e., the $L2$ loss) but is more tolerant of abnormal predictions [55].

We divided the two study areas into 30 arc-second grids, leading to a total of 28,500 ($190 \times 150$) grids in each site. Image patches in Metro Atlanta were randomly divided into a training set (17,100 patches, 60%), a validating set (5,700 patches, 20%), and a testing set (5,700 patches, 20%). Fig. 4 presents the spatial distribution of Metro Atlanta's image patches in different sets. Image patches in Metro Dallas were used to evaluate the framework's generalization capability, considering the heterogeneous patterns of land use/land cover revealed from these two study areas.

Hyperparameters were tuned empirically based on the performance of the validating set. Adam optimizer was adopted with a learning rate initialized to 0.0001 ($\beta_1 = 0.9$, $\beta_2 = 0.999$), the batch size was set to 32, and epochs were capped at 10,000, before which all selected architectures reached a stable performance. The framework run on a desktop with two NVIDIA GTX 1080Ti GPUs, a 3.20 GHz Intel Core i7-8700

CPU, and 32 GB RAM. We implemented the framework using Tensoreflow 2.0 library with Python 3.6.5 under Windows 10, CUDA 10.1, and CUDNN 7.0 system.

## D. Evaluation metrics

We used three common quantitative indices to reveal the discrepancy between $p_i^{lg}$ (ground-truthing) and $\widehat{p_i^{lg}}$ (estimated) in the testing set: Coefficient of Determination ($R^2$), Coefficient of Efficiency (CoE), and Modified Index of Agreement (MIoA). CoE, ranging from minus infinity to 1, suggests the proportion of initial variance accounted for by a model [56]. The higher the CoE value, the better agreement a model reaches. $R^2$ and MIoA are widely adopted metrics to suggest the general goodness of fit of a model. Specifically, MIoA, less sensitive to the proportional difference compared with $R^2$, is a modified version of the original Index of Agreement (IoA) proposed by Willmott et al. [57]. With a range from 0 to 1, MIoA with a higher value indicates better agreement. The calculations for the three indices follow:

$$R^2 = 1 - \frac{\sum_{i=1}^{m}(p_i^{lg} - \widehat{p_i^{lg}})^2}{\sum_{i=1}^{m}(p_i^{lg} - \overline{p_i^{lg}})^2} \quad (5)$$

$$COE = 1 - \frac{\sum_{i=1}^{m}(p_i^{lg} - \widehat{p_i^{lg}})^2}{\sum_{i=1}^{m}(p_i^{lg} - \overline{p_i^{lg}})^2} \quad (6)$$

$$MIoA = 1 - \frac{\sum_{i=1}^{m}\left|p_i^{lg} - \widehat{p_i^{lg}}\right|}{\sum_{i=1}^{m}\left(\left|\widehat{p_i^{lg}} - \overline{p_i^{lg}}\right| + \left|p_i^{lg} - \overline{p_i^{lg}}\right|\right)} \quad (7)$$

where $m$ denotes the number of testing patches and $\overline{p_i^{lg}}$ denotes the average value of $p_i^{lg}$, i.e., $\overline{p_i^{lg}} = \frac{\sum_{i=1}^{m} p_i^{lg}}{m}$.

# Atlanta

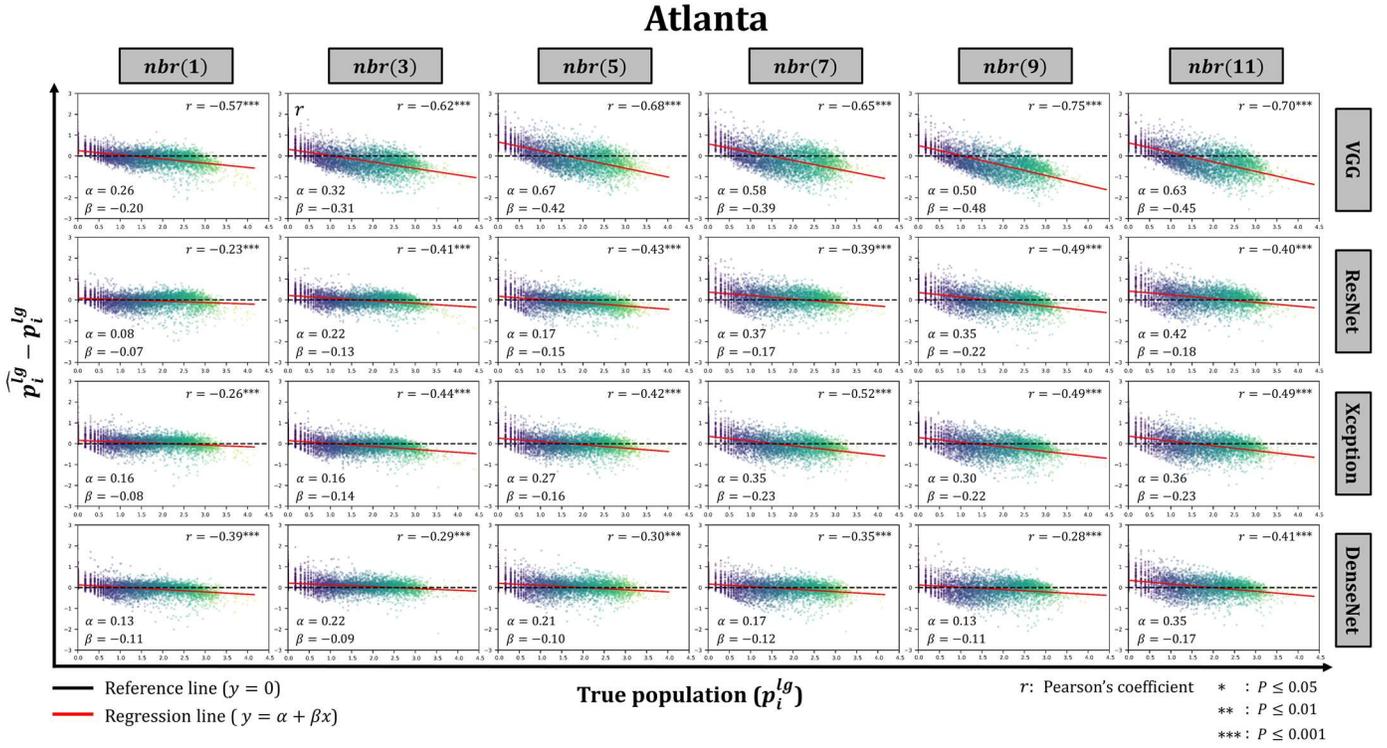

| Reference line ($y = 0$) | $r$: Pearson's coefficient | * : $P \leq 0.05$ |
| Regression line ($y = \alpha + \beta x$) | | ** : $P \leq 0.01$ |
| | | *** : $P \leq 0.001$ |

**Fig. 6.** *Scatterplot of Metro Atlanta testing patches between the true population ($p_i^{lg}$) and the difference in estimated population and true population ($\widehat{p_i^{lg}} - p_i^{lg}$) from VGG, ResNet, Xception, and DenseNet, under different neighboring scenarios.*

## V. RESULTS

### A. Model performances of testing patches in Metro Atlanta

After the model training process, we applied well-trained models to the testing patches (5,700) in Metro Atlanta, aiming to reveal the discrepancies of model performances, investigate the influence of neighboring sizes, and identify the potential systematic biases. The model performances of testing patches in Metro Atlanta under different neighborhood scenarios can be found in Table 1. In general, DenseNet stands out, as it outperforms the other three models in all evaluation metrics under all selected neighboring sizes. In comparison, VGG has the worst performance, evidenced by its lowest values in evaluating metrics among the four selected models. ResNet and Xception have a similar performance lying between DenseNet and VGG. The scatterplots of the true population ($p_i^{lg}$) and the estimated population ($\widehat{p_i^{lg}}$) also confirm this conclusion, as the scatterplot for DenseNet presents a more clustered pattern in each neighboring scenario compared to the other three models. In comparison, $p_i^{lg}$ and $\widehat{p_i^{lg}}$ are scatteredly distributed from the 1:1 reference line in VGG, suggesting its poor predicting performance.

As for neighboring scenarios, the increase of neighboring sizes leads to reduced population estimation performance, which is found universal for all four selected models in all evaluating metrics (Table 1). For instance, the $R^2$ for DenseNet with $nbr(1)$ is 0.915, which is gradually reduced when increasing neighboring patches: 0.904 with $nbr(3)$, 0.874 with

$nbr(5)$, 0.853 with $nbr(7)$, 0.836 with $nbr(9)$, and 0.811 with $nbr(11)$. Such performance reduction in DenseNet with an increased neighboring size is also supported by *CoE* and *MIoA* metrics: *CoE* gradually reduces from 0.901 to 0.773 and *MIoA* from 0.877 to 0.791, when neighboring size increases from 1 to 11. The scatterplots validate this claim, as the $p_i^{lg}$ and $\widehat{p_i^{lg}}$ in all selected models present a more scattered distribution pattern when more neighboring patches are considered (Fig. 5), suggesting that increasing neighboring sizes, to a certain degree, confuses models, leading to a negative impact on model prediction. The above results contradict the findings by Xing et al. [28], who found an improved predicting performance can be achieved with the consideration of more neighboring patches. We assume that such a phenomenon can be attributed to the heterogeneous nature of population distribution and the diminishing proportion of information in the central patch resulting from the requirement of a fixed input size for deep learning models.

To explore whether systematic biases exist in population estimation from remote sensing images via deep learning models, we investigated the relationship between the true population, i.e., $p_i^{lg}$, and the difference in estimated population and true population, i.e., $\widehat{p_i^{lg}} - p_i^{lg}$. Fig. 6 presents the scatterplots of $p_i^{lg}$ and $\widehat{p_i^{lg}} - p_i^{lg}$ from VGG, ResNet, Xception, and DenseNet, under different neighboring scenarios. The results reveal a notable, universal bias, evidenced by the negative slope ($\beta$) and negative Pearson's $r$ values significant at 0.001 significance level from all models under all

## Dallas

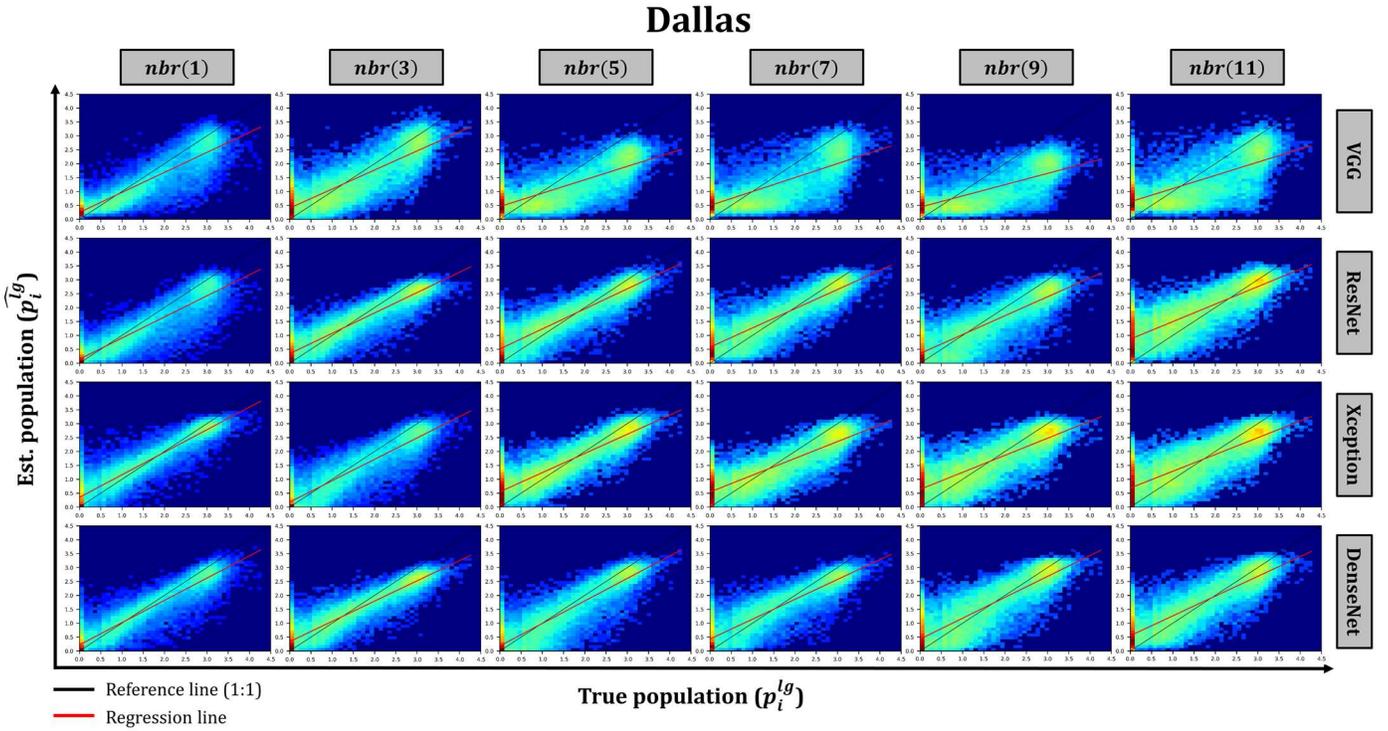

**Fig. 7.** *Scatterplot of all Metro Dallas patches between the true population ($p_i^{lg}$) and the estimated population $\widehat{(p_i^{lg})}$ from VGG, ResNet, Xception, and DenseNet, under different neighboring scenarios.*

neighboring scenarios. The above results demonstrate that all selected deep learning models tend to overestimate sparsely populated image patches and underestimate densely populated image patches, regardless of neighboring sizes. The choice of neighboring sizes plays a trivial role in the strength of such bias, as Pearson's $r$ and $\beta$ fluctuate with the increase of neighboring sizes. Among all the neighboring scenarios from $nbr(1)$ to $nbr(11)$, VGG, ResNet, and Xception have the least bias with $nbr(1)$, i.e., the central patch. For DenseNet, however, the least bias occurs with $nbr(3)$, i.e., the central patch with its eight nearby neighboring patches. The identified systematic bias can be partly explained by the intrinsic limitation of multispectral remote sensing images: lacking the vertical observation, which presumably leads to the underestimation of holding capacity (volume) of high-rise buildings in densely populated areas.

### B. Model performances of testing patches in Metro Atlanta

Given the observable discrepancies in distribution patterns of land use and land cover in Metro Dallas compared with Metro Atlanta, we evaluated the generalization capability by applying models trained using patches from Metro Atlanta to image patches in Metro Dallas. We aimed to investigate whether similar patterns still hold in Metro Dallas regarding model performances, the impact of neighboring sizes, and the identified systematic bias.

In general, DenseNet still outperforms the other three models in all evaluation metrics under all neighboring scenarios, while VGG has the worst performance (Table 2). Comparing Table 1 and Table 2, there exist overall performance reductions, mostly slight ones, for all models when they are applied to a different metroplex. For instance, the *MIoA* for DenseNet with $nbr(1)$ in Metro Atlanta is 0.877, but it is reduced to 0.838 in Metro Dallas. Similar patterns can be found for *MIoA* in DenseNet under other neighboring scenarios: from 0.867 to 0.808 with $nbr(3)$, from 0.841 to 0.793 with $nbr(5)$, from 0.831 to 0.768 with $nbr(7)$, from 0.814 to 0.741 with $nbr(9)$, and from 0.791 to 0.710 with $nbr(11)$. The above findings can be confirmed by comparing Fig. 5 and Fig. 7, where scatterplots of $p_i^{lg}$ and $\widehat{p_i^{lg}}$ are presented. Evidently, more scattered distribution of $p_i^{lg}$ and $\widehat{p_i^{lg}}$ along the 1:1 reference line can be found in Metro Dallas (Fig. 7) comparing to the corresponding ones in Metro Atlanta. The above results suggest that models trained in Metro Atlanta can be generalized to Metro Dallas. However, the differences between these two metropolitan areas pose certain challenges for all models in population estimation.

As for neighboring scenarios, the pattern identified from the testing patches in Metro Atlanta still holds: increased neighboring sizes lead to universal reduced population estimation performances for all four selected models, evidenced by gradually reduced values in all evaluating metrics (Table 2) and scattered distribution of $p_i^{lg}$ and $\widehat{p_i^{lg}}$ (Fig. 7) when neighboring scenarios move from $nbr(1)$ to $nbr(11)$.

The relationship between the $p_i^{lg}$ and $\widehat{p_i^{lg}} - p_i^{lg}$ in Dallas reveals that the same bias exists in Metro Dallas (Fig. 8) and the bias is, to some extent, intensified (Fig. 6). For instance, the *Pearson's $r$* values between $\widehat{p_i^{lg}} - p_i^{lg}$ and $p_i^{lg}$ for DenseNet

# Dallas

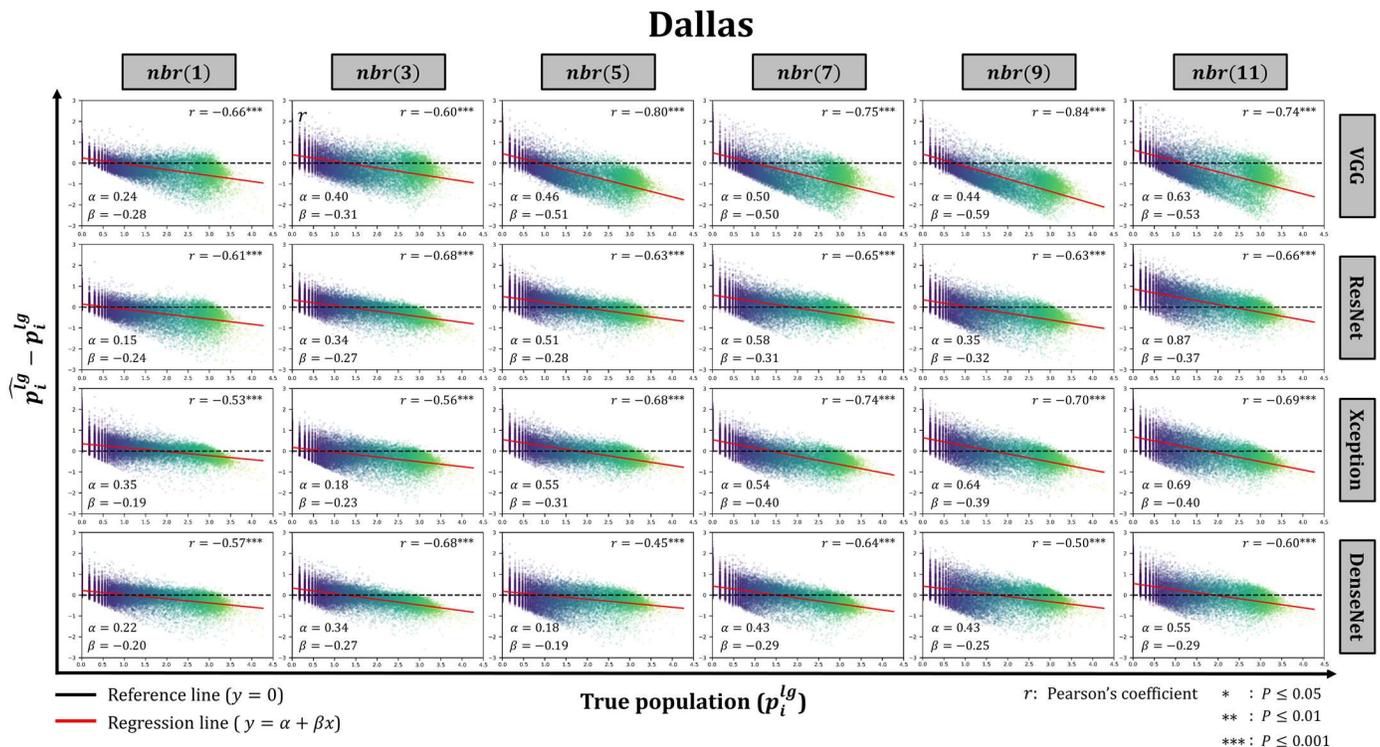

**Fig. 8.** *Scatterplot of all Metro Dallas patches between the true population $(p_i^{lg})$ and the difference in estimated population and true population $(\overline{p_i^{lg}} - p_i^{lg})$ from VGG, ResNet, Xception, and DenseNet, under different neighboring scenarios.*

with $nbr(1)$ in Metro Atlanta and Metro Dallas are -0.39 and -0.57, respectively, suggesting a strengthened negative correlation. Similar intensified biases can be found for DenseNet under other neighboring scenarios: from -0.29 to -0.68 with $nbr(3)$, from -0.30 to -0.45 with $nbr(5)$, from -0.35 to -0.64 with $nbr(7)$, from -0.28 to -0.50 with $nbr(9)$, and from -0.41 to -0.60 with $nbr(11)$. Other models that include VGG, ResNet, and Xception, also experience intensified systematic biases under varying neighboring scenarios when estimating population using image patches in Metro Dallas. The intensified bias (underestimating densely populated image patches) presumably results from the more centralized urban pattern and the denser urban core of Metro Dallas compared with Metro Atlanta (Fig. 1).

Given that all models achieve the best performance under the neighboring scenario of $nbr(1)$, we present the estimated population distribution in Metro Dallas with $nbr(1)$ via DenseNet (Fig. 9b), ResNet (Fig. 9c), Xception (Fig. 9d), VGG (Fig. 9e), as well as ground-truthing population distribution (Fig. 9a). In general, estimated population distributions from all models present a considerably similar pattern to that of the true population distribution, suggesting their overall capability of capturing the heterogeneity of population distribution from remote sensing images. However, The issue of population underestimation in urban areas is evident for all models, especially in the urban core (highlighted by the purple rectangle), where densely populated grids (dark red grids) are markedly underestimated. Our investigation reveals that the underestimated grids in the urban core contain many high-rise buildings (Fig. 9). As remote sensing imagery (Sentinel-2 images in this study) fails to obtain information in the vertical dimension, the holding capacity of these high-rise buildings is significantly underestimated, responsible for the systematic underestimation of population distribution in this region from all deep learning models. The estimated population from the best model, i.e., DenseNet, under different neighboring scenarios are presented in Fig. 10. It can be observed that increased neighboring sizes lead to blurring estimations, which is expected as the consideration of nearby neighbors essentially applies a "filter" by downplaying the importance of the central image patch, thus diminishing the grid-level heterogeneity. We can also conclude that smaller neighboring sizes are able to retain fine details in sparsely populated areas (see areas highlighted by red rectangles in Fig. 10).

## VI. Discussion

Obtaining fine-grained population distribution is of great importance to a variety of fields that demand such spatial knowledge. The rapid development of remote sensing techniques provides rich, large-coverage, and high-temporal information of the ground, which can be coupled with the emerging deep learning approaches that enable latent features and hidden geographical patterns to be extracted. In this study, we establish an end-to-end framework to evaluate the performances of four popular deep learning architectures, VGG, ResNet, Xception, and DenseNet, in estimating population distribution directly from remote sensing image patches via transfer learning and fine-tuning techniques. In addition, we conduct a thorough investigation on the neighboring effects and the potential systematic biases in population estimation using remote sensing images via deep learning approaches.

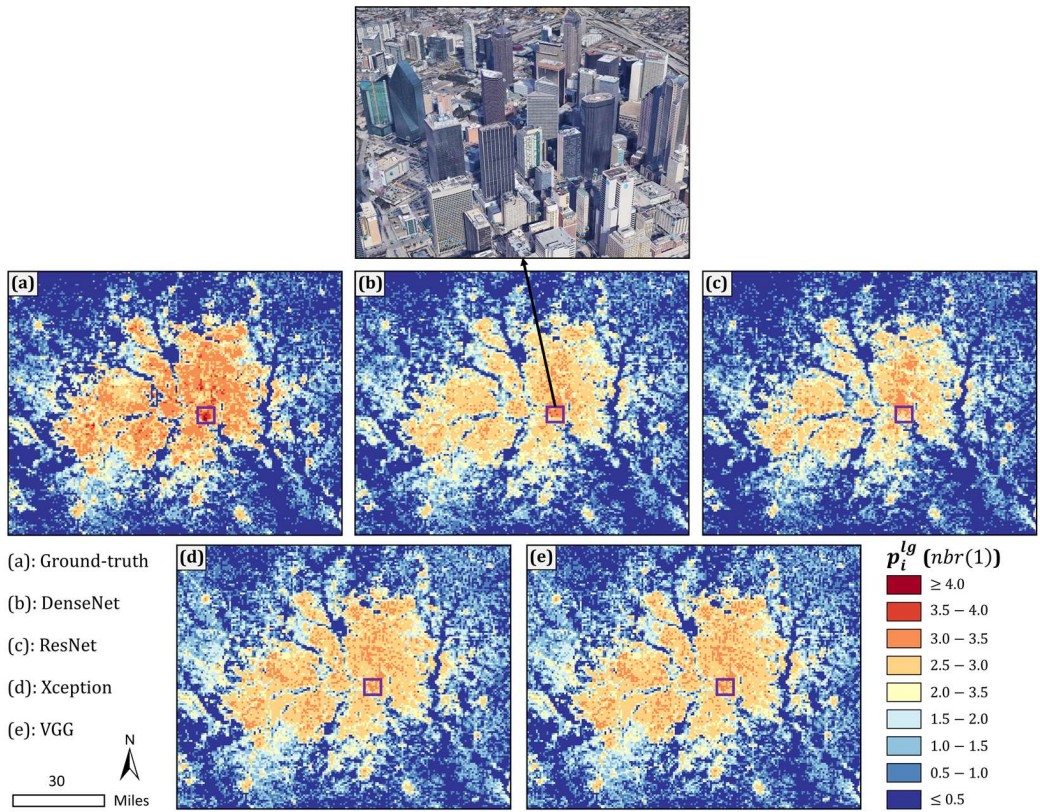

**Fig. 9.** *Ground-truthing population (a) and population estimated via DenseNet (b), ResNet (c), Xception (d), and VGG (e) under the neighboring scenario of $nbr(1)$. The 3-D imagery covered by the purple rectangle was derived from Google Map.*

Our results reveal that in all evaluation metrics under all selected neighboring sizes, DenseNet outperforms the other three models, while VGG has the worst performances. ResNet and Xception have similar performance, lying between DenseNet and VGG. We believe the superior performance of DenseNet can be attributed to its architecture design. Layers in DenseNet receive all preceding layers as input, thus creating diversified features with richer patterns and facilitating the acquisition of "collective knowledge". These diversified features and patterns assist in capturing the hidden mapping between image patches and gridded population, leading to more accurate population estimation. In addition, thanks to the channel-wise concatenation throughout its Dense Block, DenseNet maintains both low- and high-level complexity features (unlike the standard stacks of ConvNets that use mostly high-level features). The capability of DenseNet in handling multi-level features contributes to the better summarization of the heterogeneity in population distribution. In comparison, the poor performance of VGG is expected, as its deep architecture formed by plainly stacking simple convolutional operations fails to promote multi-level complexity and diversity.

In terms of the neighboring effect, our results suggest that the increase of neighboring sizes leads to reduced population estimation performance, which is found universal for all four selected models, in all evaluating metrics, and in both Metro Atlanta and Metro Dallas, contradicting a recent study by Xing et al. [28], who found $3 \times 3$ neighboring scenario outperformed $1 \times 1$ via ResNet architecture. Two possible reasons are responsible for the above findings. First, numerous studies have proved that population, as a fundamental agent in urban and suburban ecosystems, is distributed with great heterogeneity [58], [59]. For instance, a densely populated image patch can be surrounded by uninhabited patches, such as green space and water bodies. Such neighboring information does not necessarily benefit population estimation in the central patch but introduces certain uncertainty (noises) to the model prediction. Second, for the selected four deep learning models, as well as for most existing deep learning models, a fixed input shape of images is required. Considering neighboring patches unavoidably leads to a diminishing proportion of information in the central patch of an input image to the model. For instance, the proportion of information contained in the central patch with a neighboring scenario of $nbr(1)$, i.e., $1 \times 1$, is 100%. This proportion reduces to 11.11% with a neighboring scenario of $nbr(3)$, i.e., $3 \times 3$, and to 0.83% with $nbr(11)$, i.e., $11 \times 11$. Such diminishing proportion of information in the central patch with the increase of neighboring consideration poses challenges for models to predict the population residing in the central patch.

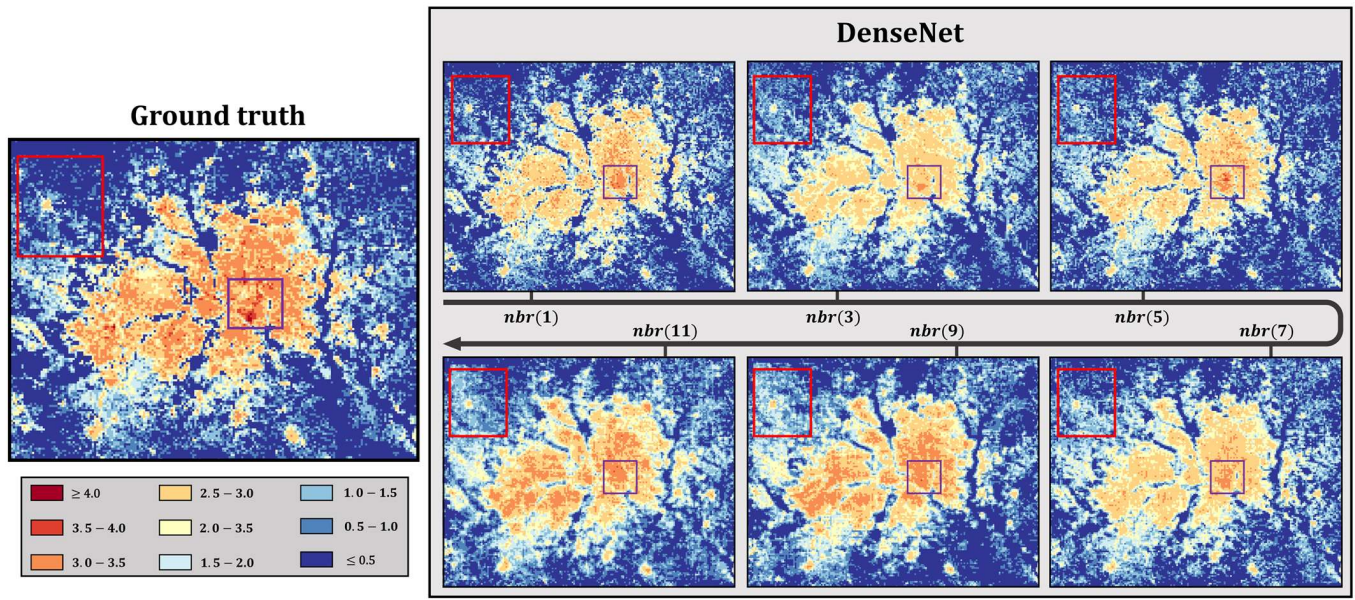

Fig. 10. Ground-truthing population and population estimated via DenseNet under different neighboring scenarios: $nbr(1)$, $nbr(3)$, $nbr(5)$, $nbr(7)$, $nbr(9)$, $nbr(11)$.

TABLE I
MODEL PERFORMANCES OF TESTING PATCHES IN METRO ATLANTA UNDER
DIFFERENT NEIGHBORHOOD SCENARIOS.

| Evaluating metrics | VGG | ResNet | Xception | DenseNet |
|---|---|---|---|---|
| $R^2$ (0~1) | | | | |
| $nbr(1)$ | 0.875 | 0.906 | 0.911 | 0.915 |
| $nbr(3)$ | 0.743 | 0.905 | 0.894 | 0.904 |
| $nbr(5)$ | 0.617 | 0.866 | 0.856 | 0.874 |
| $nbr(7)$ | 0.606 | 0.807 | 0.808 | 0.853 |
| $nbr(9)$ | 0.546 | 0.803 | 0.788 | 0.836 |
| $nbr(11)$ | 0.534 | 0.773 | 0.760 | 0.811 |
| CoE (−∞~1) | | | | |
| $nbr(1)$ | 0.827 | 0.902 | 0.905 | 0.901 |
| $nbr(3)$ | 0.590 | 0.887 | 0.870 | 0.895 |
| $nbr(5)$ | 0.371 | 0.837 | 0.838 | 0.863 |
| $nbr(7)$ | 0.350 | 0.771 | 0.747 | 0.857 |
| $nbr(9)$ | 0.130 | 0.741 | 0.723 | 0.825 |
| $nbr(11)$ | 0.105 | 0.730 | 0.707 | 0.773 |
| MIoA (0~1) | | | | |
| $nbr(1)$ | 0.830 | 0.871 | 0.865 | 0.877 |
| $nbr(3)$ | 0.738 | 0.862 | 0.852 | 0.867 |
| $nbr(5)$ | 0.672 | 0.837 | 0.832 | 0.841 |
| $nbr(7)$ | 0.666 | 0.801 | 0.792 | 0.831 |
| $nbr(9)$ | 0.617 | 0.791 | 0.785 | 0.814 |
| $nbr(11)$ | 0.589 | 0.781 | 0.769 | 0.791 |

TABLE II
MODEL PERFORMANCES IN METRO DALLAS UNDER DIFFERENT
NEIGHBORHOOD SCENARIOS.

| Evaluating metrics | VGG | ResNet | Xception | DenseNet |
|---|---|---|---|---|
| $R^2$ (0~1) | | | | |
| $nbr(1)$ | 0.817 | 0.830 | 0.835 | 0.878 |
| $nbr(3)$ | 0.732 | 0.841 | 0.820 | 0.841 |
| $nbr(5)$ | 0.563 | 0.778 | 0.756 | 0.802 |
| $nbr(7)$ | 0.546 | 0.742 | 0.711 | 0.791 |
| $nbr(9)$ | 0.476 | 0.742 | 0.654 | 0.744 |
| $nbr(11)$ | 0.452 | 0.674 | 0.624 | 0.722 |
| CoE (−∞~1) | | | | |
| $nbr(1)$ | 0.702 | 0.748 | 0.808 | 0.831 |
| $nbr(3)$ | 0.574 | 0.748 | 0.747 | 0.765 |
| $nbr(5)$ | 0.210 | 0.653 | 0.626 | 0.762 |
| $nbr(7)$ | 0.150 | 0.572 | 0.510 | 0.667 |
| $nbr(9)$ | 0.072 | 0.532 | 0.427 | 0.624 |
| $nbr(11)$ | −0.312 | 0.436 | 0.396 | 0.601 |
| MIoA (0~1) | | | | |
| $nbr(1)$ | 0.785 | 0.812 | 0.825 | 0.838 |
| $nbr(3)$ | 0.732 | 0.805 | 0.801 | 0.808 |
| $nbr(5)$ | 0.621 | 0.776 | 0.767 | 0.793 |
| $nbr(7)$ | 0.612 | 0.749 | 0.714 | 0.768 |
| $nbr(9)$ | 0.523 | 0.731 | 0.685 | 0.741 |
| $nbr(11)$ | 0.501 | 0.679 | 0.648 | 0.710 |

In terms of biases, our results reveal a notable, universal bias for all models under all neighboring scenarios, indicating that all selected deep learning models tend to overestimate sparsely populated image patches and underestimate densely populated image patches, regardless of neighboring sizes. This systematic bias can be attributed to the limitation of multispectral remote sensing sensors that lack the ability to obtain information in the vertical dimension. In densely populated urban centers, there exist a large number of high-rise buildings whose holding capacity can not be properly measured without the vertical information (i.e., building height), as they tend to hold an unexpectedly larger population than their building footprint size has suggested. Efforts have been made to extract the vertical dimension of buildings from LiDAR (Light Detection and

Ranging), aerial photogrammetry, and SAR (Synthetic Aperture Radar) [60], [61]. Future studies need to incorporate the building height information to better estimate building volume, potentially leading to better population estimation in densely populated urban areas.

Finally, we need to acknowledge several limitations of this study. First, we only compared the performances of four widely-adopted deep learning architectures with diverging concepts in design: VGG, ResNet, Xception, and DenseNet. Despite their popularity, future efforts are needed to explore the performances of other deep learning architectures in population estimation from remote sensing images. Second, we chose to estimate population distribution at 30 arc-second grids (approximately 1 km at the equator), a spatial resolution

TABLE III
CORRELATION BETWEEN THE $p_i^{lg}$ AND $\widehat{p_i^{lg}} - p_i^{lg}$ IN METRO ATLANTA AND METRO DALLAS UNDER DIFFERENT NEIGHBORING SCENARIOS.

| | VGG | | ResNet | | Xception | | DenseNet | |
|---|---|---|---|---|---|---|---|---|
| | Atlanta | Dallas | Atlanta | Dallas | Atlanta | Dallas | Atlanta | Dallas |
| **Pearson's $r$** | | | | | | | | |
| $nbr(1)$ | -0.57 | -0.66 | -0.23 | -0.61 | -0.26 | -0.53 | -0.39 | -0.57 |
| $nbr(3)$ | -0.62 | -0.60 | -0.41 | -0.68 | -0.44 | -0.56 | -0.29 | -0.68 |
| $nbr(5)$ | -0.68 | -0.80 | -0.43 | -0.63 | -0.42 | -0.68 | -0.30 | -0.45 |
| $nbr(7)$ | -0.65 | -0.75 | -0.39 | -0.65 | -0.52 | -0.74 | -0.35 | -0.64 |
| $nbr(9)$ | -0.75 | -0.84 | -0.49 | -0.63 | -0.49 | -0.70 | -0.28 | -0.50 |
| $nbr(11)$ | -0.70 | -0.74 | -0.40 | -0.66 | -0.49 | -0.69 | -0.41 | -0.60 |
| **$\alpha$** | | | | | | | | |
| $nbr(1)$ | 0.26 | 0.24 | 0.08 | 0.15 | 0.16 | 0.35 | 0.13 | 0.22 |
| $nbr(3)$ | 0.32 | 0.40 | 0.22 | 0.34 | 0.16 | 0.18 | 0.22 | 0.34 |
| $nbr(5)$ | 0.67 | 0.46 | 0.17 | 0.51 | 0.27 | 0.55 | 0.21 | 0.18 |
| $nbr(7)$ | 0.58 | 0.50 | 0.37 | 0.58 | 0.35 | 0.54 | 0.17 | 0.43 |
| $nbr(9)$ | 0.50 | 0.44 | 0.35 | 0.35 | 0.30 | 0.64 | 0.13 | 0.43 |
| $nbr(11)$ | 0.63 | 0.63 | 0.42 | 0.87 | 0.36 | 0.69 | 0.35 | 0.55 |
| **$\beta$** | | | | | | | | |
| $nbr(1)$ | -0.20 | -0.28 | -0.07 | -0.24 | -0.08 | -0.19 | -0.11 | -0.20 |
| $nbr(3)$ | -0.31 | -0.31 | -0.13 | -0.27 | -0.14 | -0.23 | -0.09 | -0.27 |
| $nbr(5)$ | -0.42 | -0.51 | -0.15 | -0.28 | -0.16 | -0.31 | -0.10 | -0.19 |
| $nbr(7)$ | -0.39 | -0.50 | -0.17 | -0.31 | -0.23 | -0.40 | -0.12 | -0.29 |
| $nbr(9)$ | -0.48 | -0.59 | -0.22 | -0.32 | -0.22 | -0.39 | -0.11 | -0.25 |
| $nbr(11)$ | -0.45 | -0.53 | -0.18 | -0.37 | -0.23 | -0.40 | -0.17 | -0.29 |

All Pearson's $r$ values are significant at a 0.001 significance level.

adopted by many studies similar to this work [27]–[29]. However, we acknowledge that the heterogeneity of population distribution tends to vary in different scales. Thus, the conclusion in this study may not hold for population estimation at a different geographical scale (e.g., in sub-km population estimation). Future studies are needed to investigate model performances, neighboring effects, and potential biases by adopting a multi-scale estimation framework. Third, we trained (17,100 patches), validated (5,700 patches), and tested (5,600 patches) deep learning models in Metro Atlanta and further applied them to Metro Dallas (28,500 patches) to evaluate their overall generalizability. As the performance of deep learning models largely depends on the size of training samples, we encourage future studies to involve more training samples from multiple study sites for better generalization capability.

## VII. Conclusion

The coupling of remote sensing imagery and deep learning algorithms undoubtedly establishes a new venue that potentially advances traditional population modeling. This study marks the first attempt to cross-compare performances of popular state-of-the-art deep learning models in estimating population distribution from remote sensing images, investigate the contribution of neighboring effect, and explore the potential systematic population estimation biases.

The results suggest that DenseNet outperforms the other three models, while VGG has the worst performances in all evaluating metrics under all selected neighboring scenarios. ResNet and Xception have similar performance, lying between DenseNet and VGG. The superior performance of DenseNet is presumably due to its architectural design that facilitates diversified features and patterns for capturing hidden mapping between image patches and gridded population. As for the neighboring effect, our results indicate that the increase of neighboring sizes leads to reduced population estimation performance, which is found universal for all four selected models, in all evaluating metrics, and in both Metro Atlanta and Metro Dallas, which contradicts a recent study by Xing et al. [28]. Such a phenomenon can be attributed to the heterogeneous nature of population distribution and the diminishing proportion of information in the central patch resulting from the requirement of a fixed input size for deep learning models. In addition, there exists a notable, universal bias that all selected deep learning models tend to overestimate sparsely populated image patches and underestimate densely populated image patches, regardless of neighboring sizes. This systematic bias can be explained by the intrinsic limitation of multispectral remote sensing images: lacking the vertical observation, thus leading to underestimating the holding capacity of high-rise buildings in densely populated urban cores. The methodological, experimental, and contextual knowledge this study provides is expected to benefit a wide range of future studies that estimate population distribution via remote sensing imagery.

## References

[1] S. I. Hay, A. M. Noor, A. Nelson, and A. J. Tatem, "The accuracy of human population maps for public health application," *Trop. Med. Int. Health*, vol. 10, no. 10, pp. 1073–1086, 2005.

[2] K. L. Frohlich and L. Potvin, "Transcending the known in public health practice: The inequality paradox: The population approach and vulnerable populations," Am. J. Public Health, vol. 98, no. 2, pp. 216–221, 2008.

[3] A. J. Tatem, A. M. Noor, and S. I. Hay, "Defining approaches to settlement mapping for public health management in Kenya using medium spatial resolution satellite imagery," Remote Sens. Environ., vol. 93, no. 1–2, pp. 42–52, 2004.

[4] M. Langford, G. Higgs, J. Radcliffe, and S. White, "Urban population distribution models and service accessibility estimation," Comput. Environ. Urban Syst., vol. 32, no. 1, pp. 66–80, 2008.

[5] J. A. Maantay, A. R. Maroko, and C. Herrmann, "Mapping population distribution in the urban environment: The cadastral-based expert dasymetric system (CEDS)," Cartogr. Geogr. Inf. Sci., vol. 34, no. 2, pp. 77–102, 2007.

[6] C. Kang, Y. Liu, X. Ma, and L. Wu, "Towards estimating urban population distributions from mobile call data," J. Urban Technol., vol. 19, no. 4, pp. 3–21, 2012.

[7] P. Tenerelli, J. F. Gallego, and D. Ehrlich, "Population density modelling in support of disaster risk assessment," Int. J. Disaster Risk Reduct., vol. 13, pp. 334–341, 2015.

[8] X. Song, Q. Zhang, Y. Sekimoto, T. Horanont, S. Ueyama, and R. Shibasaki, "Modeling and probabilistic reasoning of population evacuation during large-scale disaster," in Proceedings of the 19th ACM SIGKDD international conference on Knowledge discovery and data mining - KDD '13, 2013.

[9] C. Linard, M. Gilbert, R. W. Snow, A. M. Noor, and A. J. Tatem, "Population distribution, settlement patterns and accessibility across Africa in 2010," PLoS One, vol. 7, no. 2, p. e31743, 2012.

[10] D. Murakami and Y. Yamagata, "Estimation of gridded population and GDP scenarios with spatially explicit statistical downscaling," Sustainability, vol. 11, no. 7, p. 2106, 2019.

[11] J. Shen, "Internal migration and regional population dynamics in China," Prog. Plann., vol. 45, no. 3, pp. 123–188, 1996.

[12] C. Tacoli, "Crisis or adaptation? Migration and climate change in a context of high mobility," Environ. Urban., vol. 21, no. 2, pp. 513–525, 2009.

[13] X. Huang, C. Wang, Z. Li, and H. Ning, "A 100 m population grid in the CONUS by disaggregating census data with open-source Microsoft building footprints," Big earth data, pp. 1–22, 2020.

[14] A. S. Fotheringham and D. W. S. Wong, "The modifiable areal unit problem in multivariate statistical analysis," Environ. Plan. A, vol. 23, no. 7, pp. 1025–1044, 1991.

[15] Z. Lu, J. Im, L. Quackenbush, and K. Halligan, "Population estimation based on multi-sensor data fusion," Int. J. Remote Sens., vol. 31, no. 21, pp. 5587–5604, 2010.

[16] N. A. Wardrop et al., "Spatially disaggregated population estimates in the absence of national population and housing census data," Proc. Natl. Acad. Sci. U. S. A., vol. 115, no. 14, pp. 3529–3537, 2018.

[17] X. Huang, C. Wang, and J. Lu, "Understanding spatiotemporal development of human settlement in hurricane-prone areas on U.s. atlantic and gulf coasts using nighttime remote sensing," Nat. hazards earth syst. sci. discuss., pp. 1–22, 2019.

[18] C. L. Eicher and C. A. Brewer, "Dasymetric mapping and areal interpolation: Implementation and evaluation," Cartogr. Geogr. Inf. Sci., vol. 28, no. 2, pp. 125–138, 2001.

[19] F. Batista e Silva, J. Gallego, and C. Lavalle, "A high-resolution population grid map for Europe," J. Maps, vol. 9, no. 1, pp. 16–28, 2013.

[20] F. R. Stevens, A. E. Gaughan, C. Linard, and A. J. Tatem, "Disaggregating census data for population mapping using random forests with remotely-sensed and ancillary data," PLoS One, vol. 10, no. 2, p. e0107042, 2015.

[21] D. L. Balk, U. Deichmann, G. Yetman, F. Pozzi, S. I. Hay, and A. Nelson, "Determining global population distribution: methods, applications and data," Adv. Parasitol., vol. 62, pp. 119–156, 2006.

[22] J. Mennis, "Dasymetric Mapping for Estimating Population in Small Areas: Dasymetric mapping for estimating population in small areas," Geogr. compass, vol. 3, no. 2, pp. 727–745, 2009.

[23] X. X. Zhu et al., "Deep learning in remote sensing: A comprehensive review and list of resources," IEEE Geosci. Remote Sens. Mag., vol. 5, no. 4, pp. 8–36, 2017.

[24] Q. Zou, L. Ni, T. Zhang, and Q. Wang, "Deep learning based feature selection for remote sensing scene classification," IEEE Geosci. Remote Sens. Lett., vol. 12, no. 11, pp. 2321–2325, 2015.

[25] J. E. Ball, D. T. Anderson, and C. S. Chan, "Comprehensive survey of deep learning in remote sensing: theories, tools, and challenges for the community," J. Appl. Remote Sens., vol. 11, no. 04, p. 1, 2017.

[26] Y. LeCun, Y. Bengio, and G. Hinton, "Deep learning," Nature, vol. 521, no. 7553, pp. 436–444, May 2015. doi: 10.1038/nature14539.

[27] W. Hu et al., "Mapping missing population in rural India: A deep learning approach with satellite imagery," in Proceedings of the 2019 AAAI/ACM Conference on AI, Ethics, and Society, 2019.

[28] X. Xing et al., "Mapping human activity volumes through remote sensing imagery," IEEE J. Sel. Top. Appl. Earth Obs. Remote Sens., vol. 13, pp. 5652–5668, 2020.

[29] C. Robinson, F. Hohman, and B. Dilkina, "A deep learning approach for population estimation from satellite imagery," in Proceedings of the 1st ACM SIGSPATIAL Workshop on Geospatial Humanities, 2017.

[30] P. Doupe, E. Bruzelius, J. Faghmous, and S. G. Ruchman, "Equitable development through deep learning: The case of sub-national population density estimation," in Proceedings of the 7th Annual Symposium on Computing for Development, 2016.

[31] E. Doxsey-Whitfield et al., "Taking advantage of the improved availability of census data: A first look at the gridded population of the world, version 4," Pap. Appl. Geogr., vol. 1, no. 3, pp. 226–234, 2015.

[32] "GHS-POP R2015A - Datasets - EU data portal - Europa EU." https://data.europa.eu/euodp/en/data/dataset/jrc-ghsl-ghs_pop_gpw4_globe_r2015a (accessed: Mar. 02, 2021).

[33] C. Frye, E. Nordstrand, D. J. Wright, C. Terborgh, and J. Foust, "Using classified and unclassified land cover data to estimate the footprint of human settlement," Data Sci. J., vol. 17, 2018.

[34] B. Bhaduri, E. Bright, P. Coleman, and M. L. Urban, "LandScan USA: a high-resolution geospatial and temporal modeling approach for population distribution and dynamics," GeoJournal, vol. 69, no. 1–2, pp. 103–117, 2007.

[35] S. Eichhorn, "Disaggregating population data and evaluating the accuracy of modeled high-resolution population distribution—the case study of Germany," Sustainability, vol. 12, no. 10, p. 3976, 2020.

[36] A. Voulodimos, N. Doulamis, A. Doulamis, and E. Protopapadakis, "Deep learning for computer vision: A brief review," Comput. Intell. Neurosci., vol. 2018, pp. 1–13, 2018.

[37] A. Krizhevsky, I. Sutskever, and G. E. Hinton, "ImageNet classification with deep convolutional neural networks," Commun. ACM, vol. 60, no. 6, pp. 84–90, 2017.

[38] K. Simonyan and A. Zisserman, "Very deep convolutional networks for large-scale image recognition," arXiv [cs.CV], 2014.

[39] C. Szegedy et al., "Going deeper with convolutions," in 2015 IEEE Conference on Computer Vision and Pattern Recognition (CVPR), 2015.

[40] C. Szegedy, V. Vanhoucke, S. Ioffe, J. Shlens, and Z. Wojna, "Rethinking the inception architecture for computer vision," in 2016 IEEE Conference on Computer Vision and Pattern Recognition (CVPR), 2016.

[41] C. Szegedy, S. Ioffe, V. Vanhoucke, and A. Alemi, "Inception-v4, Inception-ResNet and the impact of residual connections on learning," arXiv [cs.CV], 2016.

[42] F. Chollet, "Xception: Deep learning with depthwise separable convolutions," in 2017 IEEE Conference on Computer Vision and Pattern Recognition (CVPR), 2017.

[43] K. He, X. Zhang, S. Ren, and J. Sun, "Deep residual learning for image recognition," in 2016 IEEE Conference on Computer Vision and Pattern Recognition (CVPR), 2016.

[44] G. Huang, Z. Liu, L. Van Der Maaten, and K. Q. Weinberger, "Densely connected convolutional networks," in 2017 IEEE Conference on Computer Vision and Pattern Recognition (CVPR), 2017.

[45] A. Ahmed, K. Yu, W. Xu, Y. Gong, and E. Xing, "Training hierarchical feed-forward visual recognition models using transfer learning from pseudo-tasks," in Lecture Notes in Computer Science, Berlin, Heidelberg: Springer Berlin Heidelberg, 2008, pp. 69–82.

[46] J. Yosinski, J. Clune, Y. Bengio, and H. Lipson, "How transferable are features in deep neural networks?," arXiv [cs.LG], 2014.

[47] "Oak Ridge national laboratory," Ornl.gov. [Online]. Available: http://web.ornl.gov/. [Accessed: 02-Mar-2021].

[48] J. E. Dobson, E. A. Bright, P. R. Coleman, R. C. Durfee, and B. A. Worley. "LandScan: a global population database for estimating populations at risk," Photogramm. Eng. Remote Sensing, vol. 69, no. 7, pp. 849–857, 2000.

[49] P. P. Simarro et al., "Estimating and mapping the population at risk of sleeping sickness," PLoS Negl. Trop. Dis., vol. 6, no. 10, p. e1859, 2012.

[50] A. Smith, P. D. Bates, O. Wing, C. Sampson, N. Quinn, and J. Neal, "New estimates of flood exposure in developing countries using high-resolution population data," Nat. Commun., vol. 10, no. 1, p. 1814, 2019.

[51] "LandScan USA," Arcgis.com. [Online]. Available: https://hifld-geoplatform.opendata.arcgis.com/datasets/e431a6410145450aa56606568345765b. [Accessed: 02-Mar-2021].

[52] M. Drusch et al., "Sentinel-2: ESA's optical high-resolution mission for GMES operational services," Remote Sens. Environ., vol. 120, pp. 25–36, 2012.

[53] M. Herold, D. A. Roberts, M. E. Gardner, and P. E. Dennison, "Spectrometry for urban area remote sensing—Development and analysis of a spectral library from 350 to 2400 nm," Remote Sens. Environ., vol. 91, no. 3–4, pp. 304–319, 2004.

[54] Z. Jiang et al., "Analysis of NDVI and scaled difference vegetation index retrievals of vegetation fraction," Remote Sens. Environ., vol. 101, no. 3, pp. 366–378, 2006.

[55] "Log Ridge Hyperbolic Cosine Loss Improves Variational Auto-Encoder" https://openreview.net/forum?id=rkglvsC9Ym (accessed: Mar. 02, 2021).

[56] A. N. Mandeville, P. E. O'Connell, J. V. Sutcliffe, and J. E. Nash, "River flow forecasting through conceptual models part III - The Ray catchment

at Grendon Underwood," J. Hydrol. (Amst.), vol. 11, no. 2, pp. 109–128, 1970.

[57] C. J. Willmott et al., "Statistics for the evaluation and comparison of models," J. Geophys. Res., vol. 90, no. C5, p. 8995, 1985.

[58] D. Azar et al., "Spatial refinement of census population distribution using remotely sensed estimates of impervious surfaces in Haiti," Int. J. Remote Sens., vol. 31, no. 21, pp. 5635–5655, 2010.

[59] X. Huang, C. Wang, and Z. Li, "High-resolution population grid in the CONUS using microsoft building footprints: A feasibility study," in Proceedings of the 3rd ACM SIGSPATIAL International Workshop on Geospatial Humanities - GeoHumanities '19, 2019.

[60] C. Stal, F. Tack, P. De Maeyer, A. De Wulf, and R. Goossens, "Airborne photogrammetry and lidar for DSM extraction and 3D change detection over an urban area – a comparative study," Int. J. Remote Sens., vol. 34, no. 4, pp. 1087–1110, 2013.

[61] D. Brunner, G. Lemoine, L. Bruzzone, and H. Greidanus, "Building height retrieval from VHR SAR imagery based on an iterative simulation and matching technique," IEEE Trans. Geosci. Remote Sens., vol. 48, no. 3, pp. 1487–1504, 2010.